# On the validation of pansharpening methods

Gintautas Palubinskas

Remote Sensing Technology Institute,
German Aerospace Center (DLR),
Oberpfaffenhofen, 82234 Wessling, Germany,
E-Mail: Gintautas.Palubinskas@dlr.de

**Abstract:** Validation of the quality of pansharpening methods is a difficult task because the reference is not directly available. In the meantime, two main approaches have been established: validation in reduced resolution and original resolution. In the former approach it is still not clear how the data are to be processed to a lower resolution. Other open issues are related to the question which resolution and measures should be used. In the latter approach the main problem is how the appropriate measure should be selected. In the most comparison studies the results of both approaches do not correspond, that means in each case other methods are selected as the best ones. Thus, the developers of the new pansharpening methods still stand in the front of dilemma: how to perform a correct or appropriate comparison/evaluation/validation. It should be noted, that the third approach is possible, that is to perform the comparison of methods in a particular application with the usage of their ground truth. But this is not always possible, because usually developers are not working with applications. Moreover, it can be an additional computational load for a researcher in a particular application. In this paper some of the questions/problems raised above are approached/discussed. The following component substitution (CS) and high pass filtering (HPF) pansharpening methods with additive and multiplicative models and their enhancements such as haze correction, histogram matching, usage of spectral response functions (SRF), modulation transfer function (MTF) based lowpass filtering are investigated on remote sensing data of WorldView-2 and WorldView-4 sensors.

## 1. Introduction

Pansharpening aims to restore/estimate a high resolution multispectral image from two input images: a low resolution multispectral image (MS) and a high resolution panchromatic image (Pan). Generally, multi-resolution image fusion is not limited only to multispectral and panchromatic image pairs. For example, in so called hyper-sharpening, a low resolution image is a hyperspectral image and a high resolution image is a multispectral image. A large number of algorithms and methods to solve this type of problem were introduced during the last three decades, e.g. [1-12] with different method classifications or groupings proposed. In general, they can be divided into three large groups: component substitution (CS) methods, high pass filtering (HPF) and optimization methods based on the minimization of model error residuals [13-19, 45].

Several pansharpening improvement techniques such as panchromatic image histogram matching [20], inclusion of image formation/acquisition model into pansharpening process [2, 13, 21-25, 26], haze removal [27], MTF based low pass filtering [28, 29] widely spread during the recent decade.

Validation of the quality of pansharpening methods is a difficult task because the reference is not directly available. Two main approaches have been established: validation in reduced resolution [30] and original resolution. In the former case the reference exists, thus a lot of quality measures have been proposed [3, 31-35, 44]. In original resolution the reference is missing and quite few quality measures exist, e.g. [36], which are still not satisfactorily as shown in [37].

In this paper several enhancements of the most popular CS and HPF based pansharpening methods such as haze correction [27, 38], panchromatic band histogram matching [20, 26], spectral weights estimation used for intensity image calculation [24, 26], model-based panchromatic band correction [26] and pansharpening result histogram matching [26] are investigated for the two remote sensing satellite data: WorldView-2 and WorldView-4.

Validation approach used in this paper consists of comparing two most popular quality measures: ERGAS [39] and SAM [40] in several reduced scales. Correlation analysis is applied on the results of validation to support evaluation of comparison/experiments.

This paper presents the validation of the quality of pansharpening methods. The paper is organized as follows. In section 2 definitions used in the paper are presented. In the following sections preparation of input images and parameters (Sect. 3), pansharpening methods (Sect 4), enhancements of pansharpening methods (Sect. 5) are described. Then, in section 6 validation approach used in this paper is introduced. In section 7 the results of experiments are presented and analyzed. Finally, the paper ends with discussion, conclusions and reference sections.

## 2. Definitions

Below definitions/notations for images and their pixels are introduced which are used in this paper [26].

*2.1. Low resolution images*

$S_{lr} = \{s_{lr,k,i}, k = 1, ..., K, i = 1, ..., I\}$ – low resolution multispectral image matrix (also MS image or MS bands), $k$ – band index, $K$ - number of bands, $i$ – pixel index (for simplicity it is displayed as a one-dimensional index though in reality is a two-dimensional index), $I$- number of pixels. It is an input image for pansharpening.

$I_{lr} = \{i_{lr,i}, i = 1, ..., I\}$ - intensity image calculated from multispectral image $S_{lr}$ using weights $W = \{w_k\}$ based on sensor spectral response functions (usually available from data provider)

$$i_{lr,i} = \sum_1^K w_k \cdot s_{lr,k,i} \quad (1)$$

or in matrix form

$$I_{lr} = W \cdot S_{lr}. \quad (2)$$

$P_{lr} = \{p_{lr,i}, i = 1, ..., I\}$ – low resolution panchromatic image (Pan image) calculated from high resolution panchromatic image $P_{hr}$ (for definition see the following sub-section) using a low pass filter $F = \{f_j\}$ designed to account resolution change from high resolution to low resolution

$$p_{lr,i} = \sum_{j \in F} f_j \cdot p_{hr,j} \quad (3)$$

where $f_j$ – low pass filter values, index $j$ runs all indices inside a low pass filter $F$. $F$ can include decimation or sub-sampling operator. In matrix form

$$P_{lr} = F \cdot P_{hr}. \quad (4)$$

*2.2. High resolution images*

$S_{hr} = \{s_{hr,k,j}, k = 1, ..., K, j = 1, ..., J\}$ – high resolution multispectral image, $k$ – band index, $j$ – pixel index, $J$- number of pixels. It is the aim/result of pansharpening or an output of pansharpening.

$\tilde{S}_{hr}$ – upsampled, e.g. using cubic convolution interpolation, multispectral image in low resolution $S_{lr}$ to high resolution.

$I_{hr} = \{i_{hr,j}, j = 1, ..., J\}$ - intensity image calculated from multispectral image $S_{hr}$ using weights $W$ based on spectral response functions

$$i_{hr,j} = \sum_1^K w_k \cdot s_{hr,k,j} \quad (5)$$

or in matrix form

$$I_{hr} = W \cdot S_{hr} \quad (6)$$

$P_{hr} = \{p_{hr,j}, j = 1, ..., J\}$ – high resolution panchromatic image. It is an input for pansharpening.

## 3. Preparation of input images and parameters for pansharpening

In modern pansharpening methods the following input parameters such as sensor spectral response function (SRF) and modulation transfer function (MTF) are needed. These parameters can be acquired directly from remote sensing data providers or estimated in an image. Preparation of input images by minimizing the spatial shift between Pan and MS images is presented in the following Sect. 3.1. Estimation of SRF in input images is presented in the Sect. 3.2. Finally, MTF-based low pass filtering of Pan and MS images is described in Sections 3.3 and 3.4.

*3.1. Minimize input images shift*

It is quite often that MS and Pan images are not aligned perfectly in spatial domain and thus possible shift should be corrected for a successful pansharpening. The following approach has been proposed to minimize $S_{lr}$ and $P_{hr}$ image shift:
- upsample (e.g. using cubic convolution interpolation) $S_{lr}$ to $\tilde{S}_{hr}$
- calculate intensity $\tilde{I}_{hr}$ from $\tilde{S}_{hr}$ using (6)

- minimize the following measure $RMSE(P_{hr}, \tilde{I}_{hr}) = \sqrt{1/J \cdot \Sigma_1^J (P_{hr,j} - \tilde{I}_{hr,j})^2}$
- shift images/image if necessary

It should be noted, that if data provider weights are not available, then they can be estimated using approach proposed in Sect. 3.2.

*3.2. SRF*

Let's assume two optical sensors with a spectral sensor $S$ usually exhibiting several narrow spectral response functions (SRF) for each band and a panchromatic sensor $P$ with usually broad SRF, e.g. see Figure 1. It is obvious, that a simple averaging of spectral bands will be different radiometrically from a panchromatic band.

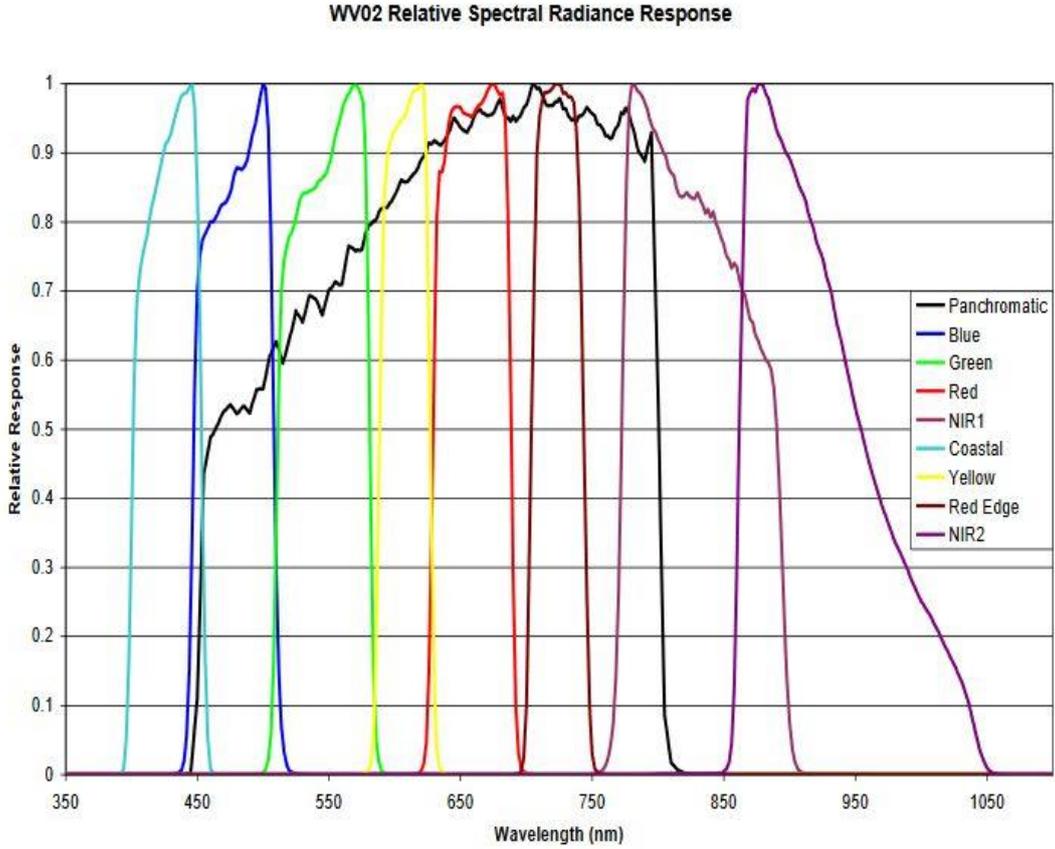

**Figure 1.** Spectral response of the WorldView-2 panchromatic and multispectral imagery [41].

For a successful fusion of data from such two sensors the energy balance between radiances/reflectances of both sensors: intensity calculated from spectral bands according (2, 6) and panchromatic image should hold, that is

$$\Sigma_1^K w_k \cdot s_{k,i} = p_i \tag{7}$$

or in vector/matrix form

$$W \cdot S = P \tag{8}$$

where $W$ are weights or SRF for individual spectral bands. It is assumed, that weights $W$ describe properties of a multispectral signal and thus are independent on the image resolution or scale.

3.2.1. Estimation of weights in low resolution image domain

Weights $\widehat{W}$ in (8) are estimated using available multispectral image in low resolution from the following equation

$$W \cdot S_{lr} = P_{lr} \tag{9}$$

e.g. by using bounded-variable least-squares (BVLS) minimization with a constraint $0 \leq w_k \leq 1$ [42].

This optimization or weights estimation is independent on its initialization values. Thus, when the provider weights are not known, then equal values can be assumed, e.g. $W_0 = 1/K$.

Here it should be noted, that other ways of estimating these weights can be used as e.g. proposed in [24] by using a linear regression algorithm. No constraint on weights can result in negative weight values, what contradicts a physical image acquisition model. Thus, this alternative approach is not considered in this paper.

3.2.2. Estimation of weights in high resolution image domain

Weights $\widehat{W}$ in (8) can be estimated using upsampled multispectral image in high resolution $\tilde{S}_{hr}$ [24] from the following equation

$$W \cdot \tilde{S}_{hr} = P_{hr} \tag{10}$$

using the same approach as in Sect. 3.2.1. It should be noted, that no advantage of this estimation has been observed, thus it is not treated in this paper.

*3.3 Low pass filtering*

Low pass filtering for downsampling of input images is needed for the validation in a reduced scale or for application of some quality measures. Two types of low pass filtering are mostly used for downsampling: Gaussian and Butterworth filters.

Low pass Gaussian filter is defined as

$$G(f) = e^{-\frac{1}{2}\left(\frac{f}{f_c}\right)^2} \tag{11}$$

where cutoff frequency is $f_c = \frac{L}{2} \cdot \frac{1}{scale}$, $L$ – filter length, scale – scale between Pan and MS resolution, e.g. usually 4.

Low pass Butterworth filter is defined as

$$B(f) = \frac{1}{1 + c\left(\frac{f}{f_c}\right)^{2n}} \tag{12}$$

where cutoff frequency is the same as for Gaussian filter, $c = 1 \text{ or } \sqrt{2}$, n is an integer number $n \geq 1$. Butterworth filter with parameter set: $c = \sqrt{2} \text{ and } n = 2$ is quite similar to Gaussian filter and is used in this paper.

It should be noted that the filters (11, 12) are defined in a spectral domain. For the use in a spatial domain they should be transformed by an inverse FFT.

*3.4. MTF based low pass filtering*

It is already broadly accepted that the low pass filtering of imagery used in pansharpening should account for the MTF values at Nyquist frequency of a particular sensor [28].

MTF values at Nyquist frequency for most popular sensors are presented in Table 1 [12].

**Table 1.** MTF values at Nyquist frequency for some of the sensors.

| Sensor | Pan | MS |
|---|---|---|
| QB | 0.15 | 0.34, 0.32, 0.30, 0.22 |
| IKONOS | 0.17 | 0.26, 0.28, 0.29, 0.28 |
| GeoEye-1, WV-4 | 0.16 | 0.23, 0.23, 0.23, 0.23 |
| WV-2 | 0.11 | 0.35, 0.35, 0.35, 0.35, 0.35, 0.35, 0.35, 0.27 |
| WV-3 | 0.14 | 0.325, 0.355, 0.36, 0.35, 0.365, 0.36, 0.335, 0.315 |
| Default | 0.15 | 0.3, 0.3, 0.3, 0.3, 0.3, 0.3, 0.3, 0.3 |

Band order for 4 bands sensors: B, G, R, NIR.

In order to incorporate the sensor's MTF value at Nyquist in the low pass filtering the cutoff frequency should be adapted in the following way: for the Gaussian filter $f_c = \frac{L}{2} \cdot \frac{1}{scale} \cdot 1 / \sqrt{-2 \ln MTF_{Nyquist}}$, $L$ – filter length, scale – scale between Pan and MS resolution, e.g. usually between panchromatic and multispectral it is 4, $MTF_{Nyquist}$ is a value from the Table 1 and for the Butterworth filter $f_c = \frac{L}{2} \cdot \frac{1}{scale} \cdot \left(\frac{c}{MTF_{Nyquist}}\right)^{\frac{1}{2n}}$.

## 4. Pansharpening methods

In this work classical CS based and HPF based pansharpening methods are analyzed, which in last decades have received various enhancements, e.g. extension of CS to any number of bands. Both groups of methods can be implemented using additive model (the following notation further CS a or HPF a is used) or multiplicative model (CS m or HPF m). For more information see following sections.

### 4.1. CS based methods

For CS (Component Substitution) based methods using additive model the pansharpening equation will look like

$$\hat{S}_{hr} = \tilde{S}_{hr} + \hat{P}_{hr} - \tilde{I}_{hr} \tag{13}$$

where $\tilde{S}_{hr}$ is an interpolated/upsampled, e.g. using bicubic convolution, version of $S_{lr}$, $\tilde{I}_{hr}$ intensity image calculated from multispectral image $\tilde{S}_{hr}$ using estimated weights $\widehat{W}$ (see Sect. 3.2), $\hat{P}_{hr}$ panchromatic image after the application of one possible enhancement presented in Sect. 5. For multiplicative model

$$\hat{S}_{hr} = \tilde{S}_{hr} \cdot \hat{P}_{hr} / \tilde{I}_{hr} \tag{14}$$

### 4.2. HPF based methods

For HPF (high pass filter) based methods using additive model the equation is the following

$$\hat{S}_{hr} = \tilde{S}_{hr} + \hat{P}_{hr} - \hat{P}_{lr} \tag{15}$$

where $\tilde{S}_{hr}$ is an interpolated, e.g. using bicubic convolution, version of $S_{lr}$, $\hat{P}_{hr}$ panchromatic image after the application of one possible enhancement presented in Sect. 5, $\hat{P}_{lr}$ is a low pass filtered version of $\hat{P}_{hr}$. For multiplicative model

$$\hat{S}_{hr} = \tilde{S}_{hr} \cdot \hat{P}_{hr} / \hat{P}_{lr} \tag{16}$$

## 5. Enhancements of pansharpening methods

Following five corrections/improvements of pansharpening methods are treated: haze correction [27], panchromatic image histogram matching [20, 26], weights estimation used for intensity image calculation [24, 26], panchromatic image correction including weight estimation [26] and finally pansharpening output/result histogram matching [26].

### 5.1. Haze correction

Recently it has been shown that the correction of the path-radiance term introduced by the atmosphere can lead to the enhancement of CS and HPF multiplicative model based pansharpening methods [27, 38]: "the path radiance, which appears as a haze in a true-color display, should be estimated and subtracted from each band before modulation and reinserted later, to restore the unbiased sharpened image" [27].

For CS multiplicative model

$$\hat{S}_{hr}(k) = \left(\tilde{S}_{hr}(k) - H(k)\right) \cdot \frac{\left(\hat{P}_{hr} - H\right)}{\left(\tilde{I}_{hr} - H\right)} + H(k) \tag{17}$$

where $H(k)$ – haze value for MS band $k$, $H$ – haze for both intensity and panchromatic image and is equal to $H = \sum_1^K \widehat{w}_k \cdot H(k)$.

For HPF multiplicative model

$$\hat{S}_{hr}(k) = \left(\tilde{S}_{hr}(k) - H(k)\right) \cdot \frac{\left(\hat{P}_{hr} - H\right)}{\left(\hat{P}_{lr} - H\right)} + H(k) \tag{18}$$

According to the authors [27, 38]: "reasonably verified in the practice, the path radiance is the minimum of the spectral radiance for each spectral band". It has been found in [29] that the modeled path radiance is well approximated by 95% of the 1-percentile of blue band, 65% of the 1-percentile of green band, 45% of the 1-percentile of red band, and 5% of the 1-percentile of NIR band for four band sensors. For other sensors the following haze estimation $H(k) = \min_j \left( \tilde{S}_{hr}(k,j) \right)$ has been used in this paper.

*5.2. Pan histogram matching*

Due to various reasons, e.g. calibration inaccuracies, image acquisition by different sensors, input images for pansharpening: low resolution multispectral image (or calculated intensity image) and high resolution panchromatic image may differ in their physical properties, e.g. radiances or reflectances depending on the processing level. For physically justified fusion, the same objects/classes in both images must exhibit very similar intensity/panchromatic values or more generally similar statistics. Similarity definition will depend on the particular application.

Thus, adjustment of Pan image to MS image or calculated intensity image is needed for a successful pansharpening. Usually histogram matching is applied. A new model-based Pan image adjustment method has been proposed in [26] and will be described in Sect 5.4. Calculation of intensity image from MS bands is treated in Sect. 3.2. Data provider initial weights or estimated weights maybe necessary as an input.

In the workflow of most pansharpening methods two possibilities are available: Pan image histogram matching before fusion or MS image histogram matching after fusion (see Sect. 5.5).

By histogram matching in general I understand adjustment of histogram or more precisely its shape of one image to the histogram of another image by using cumulative histograms. I call it further a full histogram matching. An alternative approach is to match only the first-order statistics (mean and variance) of both images, as e.g. proposed in [20]. Further this type of histogram matching will be called a simple histogram matching. It will work successful only for normal distributions.

Physically justified histogram matching is reasonable between Pan and intensity images, and original MS bands and MS bands after fusion. It should be noted, that both histogram matching approaches will work for imagery of different scales/resolutions. I have to note, that sometimes Pan and individual MS band histogram matching are used for histogram matching. Due to different spectral properties of imagery such type of histogram matching is not justified physically and thus not treated in this paper.

5.2.1. Full histogram matching

In full histogram matching a histogram of high-resolution panchromatic image $P_{hr}$ is adjusted to the histogram of intensity image $I_{lr}$ calculated from original low resolution MS image $S_{lr}$. Of course, implicitly it is assumed, that the histogram form/shape does not depend on the scale/resolution. It can be noted here, that it is possible to match Pan image to intensity image $\tilde{I}_{hr}$ calculated from up-sampled MS image $\tilde{S}_{hr}$, as it is proposed in [20]. First option, using original MS image can be preferred against interpolated/up-sampled MS image due to possible artefacts introduced by interpolation method used.

5.2.2. Simple histogram matching

Simple histogram matching denotes "the matching of the first-order statistics (mean and variance) of the Pan image to those of each MS band, or of a combination of them e.g. intensity" [20]. It should be noted, that physically it makes no sense matching Pan image to separate MS bands, thus this option is not considered in this paper.

Formulae for Pan image matching to intensity calculated from original MS bands looks like [20]

$$\hat{P}_{hr} = \left( P_{hr} - \mu_{P_{hr}} \right) \cdot \left( \sigma_{I_{lr}} / \sigma_{P_{hr}} \right) + \mu_{I_{lr}} \qquad (19)$$

in which μ and σ denote the mean and standard deviation respectively.

As already stated in the previous sub-section two options are possible: Pan adjustment to low resolution $I_{lr}$ (proposed in this paper) or Pan adjustment to high resolution $\tilde{I}_{hr}$ [20].

*5.3. Weights estimation: use updated weights for intensity calculation*

Weights needed for the intensity calculation can be acquired in two main ways: use data provider information or perform estimation directly in input image [24, 26].

*5.4. Pan correction including weights estimation*

A new model-based image adjustment method [26] performs correction of Pan image respecting physics of image formation/acquisition process, thus being superior against the image-based histogram matching.

5.4.1. Image formation/acquisition model

The idea of input image adjustment will be described below. Let's assume two optical sensors with a spectral sensor $S$ usually exhibiting several narrow spectral response functions (SRF) for each band and a panchromatic sensor $P$ with usually broad SRF, e.g. see Figure 1. It is obvious, that a simple averaging of spectral bands will be different radiometrically from a panchromatic band.

For a successful fusion of data from such two sensors the energy balance between radiances/reflectances of both sensors should hold, that is

$$\sum_1^K w_k \cdot s_{k,i} + v_i = p_i \qquad (20)$$

or in vector/matrix form

$$W \cdot S + V = P \qquad (21)$$

where $V$ is a virtual band introduced to compensate for total energy differences in different sensors and $W$ are weights for individual spectral bands. It is assumed, that weights $W$ describe properties of a multispectral signal and thus are independent on the image resolution or scale. Further, it should be noted, that a virtual band has been already introduced in [43], but without any practical solution how to estimate it.

5.4.2. Estimation of virtual band in high resolution image domain

From (20) it follows that

$$v_{lr,i} = p_{lr,i} - \sum_1^K \widehat{w}_k \cdot s_{lr,k,i} \qquad (22)$$

or in vector form

$$V_{lr} = P_{lr} - \widehat{W} \cdot S_{lr} . \qquad (23)$$

Further, this virtual band $V_{lr}$ is interpolated, e.g. using bicubic convolution, to a high resolution image $\widetilde{V}_{hr} = \{v_{hr,j}\}$ .

Finally, in order (21) to be hold, $P_{hr}$ is corrected/adjusted using the following equation:

$$\widehat{P}_{hr} = P_{hr} - \widetilde{V}_{hr} . \qquad (24)$$

This new panchromatic image $\widehat{P}_{hr}$ ensures more similar statistical radiometric properties with $I_{hr}$ and thus can be used to enhance any pansharpening method as shown in the following sections.

*5.5. Pansharpening result histogram matching*

Histogram matching as described in Sect. 5.2 can be applied to match the result of pansharpening to original MS image on per band basis [26]. Combinations of two and more simultaneous corrections are possible, but these are not treated in this paper due to the application dependency. Moreover, presently established validation approaches are still not mature as it will be shown in section 7.

## 6. Validation approach

Pansharpening methods are validated for data of two remote sensing sensors: WorldView-2 satellite exhibiting eight spectral bands (two image regions of interest (ROIs) are selected) and WorldView-4 – four spectral bands.

Validation of pansharpening is a difficult task because the reference does not exist in the original scale. One of possible approaches is to validate pansharpening in a reduced scale. In this case the reference exists and thus reference-based measures can be applied. In this paper two popular measures are used: ERGAS [39] and SAM [40]. Second approach is to perform validation in original scale. In this case the reference is not available. These exist some no-reference measures, e.g. in [36], but they are still not reliable as shown in [37]. Thus, only visual interpretation can be performed in original scale, which is quite often very subjective. Third validation approach and in my opinion the optimal one could be performed for a particular application using available ground truth (GT) data. The main drawback is that such information/knowledge is usually not available for the method developer. Thus, the first two approaches are mostly used/spread in the practice.

*6.1. Data*

WorldView-2 satellite remote sensing data (two different regions of interest ROIs selected) over the Munich city in South Germany and WorldView-4 data over Berlin city have been used in the following experiments. For summary of scene details see Table 2.

**Table 2.** Scene parameters for WorldView-2 data over Munich city and for WorldView-4 data over Berlin city.

| Parameter | Value | Value |
| --- | --- | --- |
| Sensor | WV-2 | WV-4 |
| Image date | 12-Jul-2010 | 03-Aug-2018 |
| Image time (local) | 10:30:17 | 10:34:27 |
| Mode | PAN+MS | PAN+MS |
| Product | L2A | L2A |
| Resolution PAN (m) | 0.5 | 0.3 |
| Resolution MS (m) | 2.0 | 1.2 |

For WorldView-2 data provider (DigitalGlobe) weights (spectral response functions) are available (Figure 1). In this paper they are normalized to the total energy of panchromatic band: $W_0 =$ [0.0074, 0.1106, 0.1787, 0.12076, 0.1987, 0.1363, 0.0959, 0.0002793]. For WorldView-4 data unfortunately these weights are not available, thus estimated values have been used: $W_0 =$ [0.242, 0.1025, 0.3596, 0.103].

MS images in original resolution and image size of 1024x1024 pixels are presented in Figures 2-4.

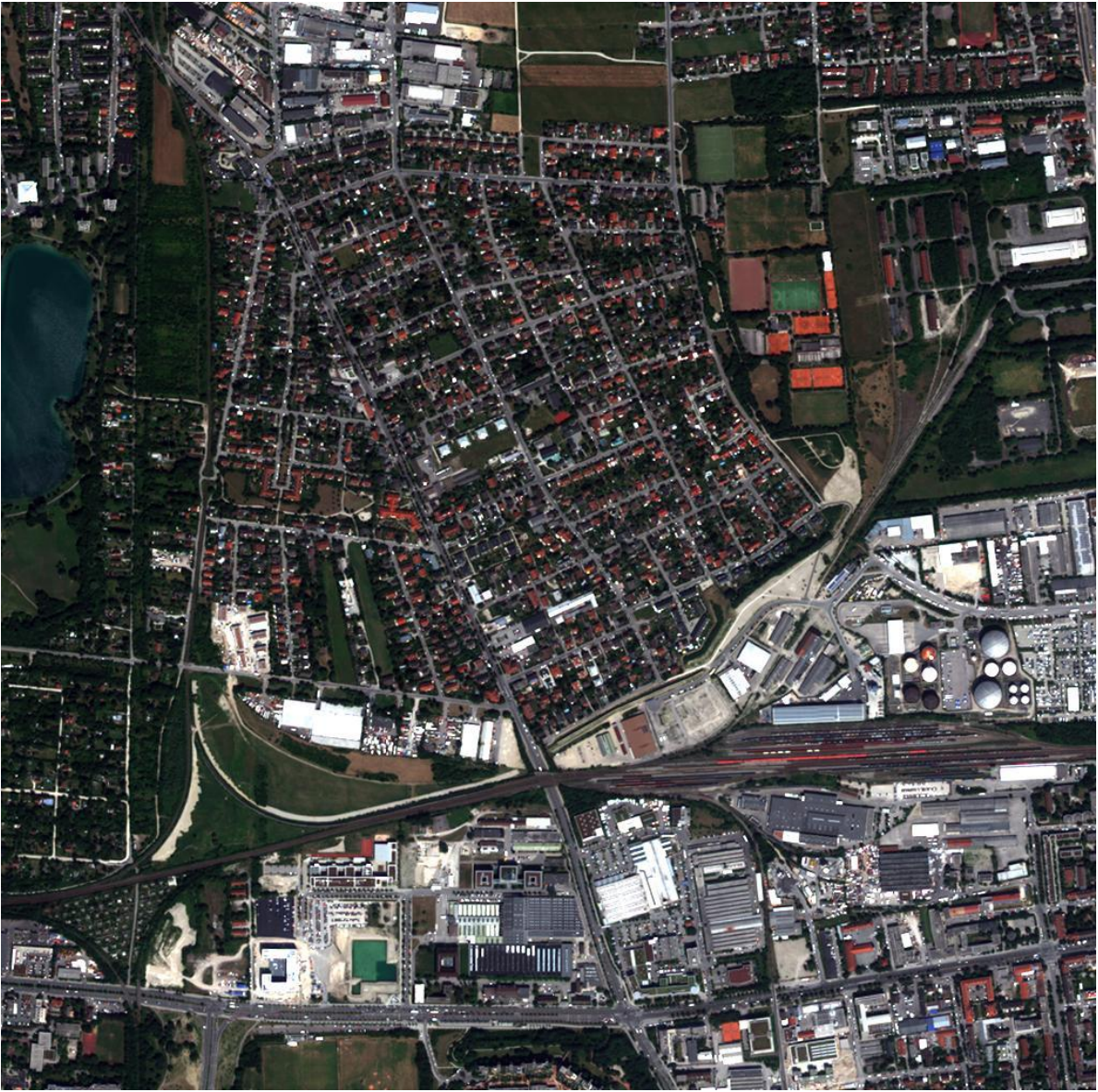

**Figure 2.** Multispectral image (bands: 5, 3, 2, size: 1024x1024) in original resolution for WV-2 ROI1.

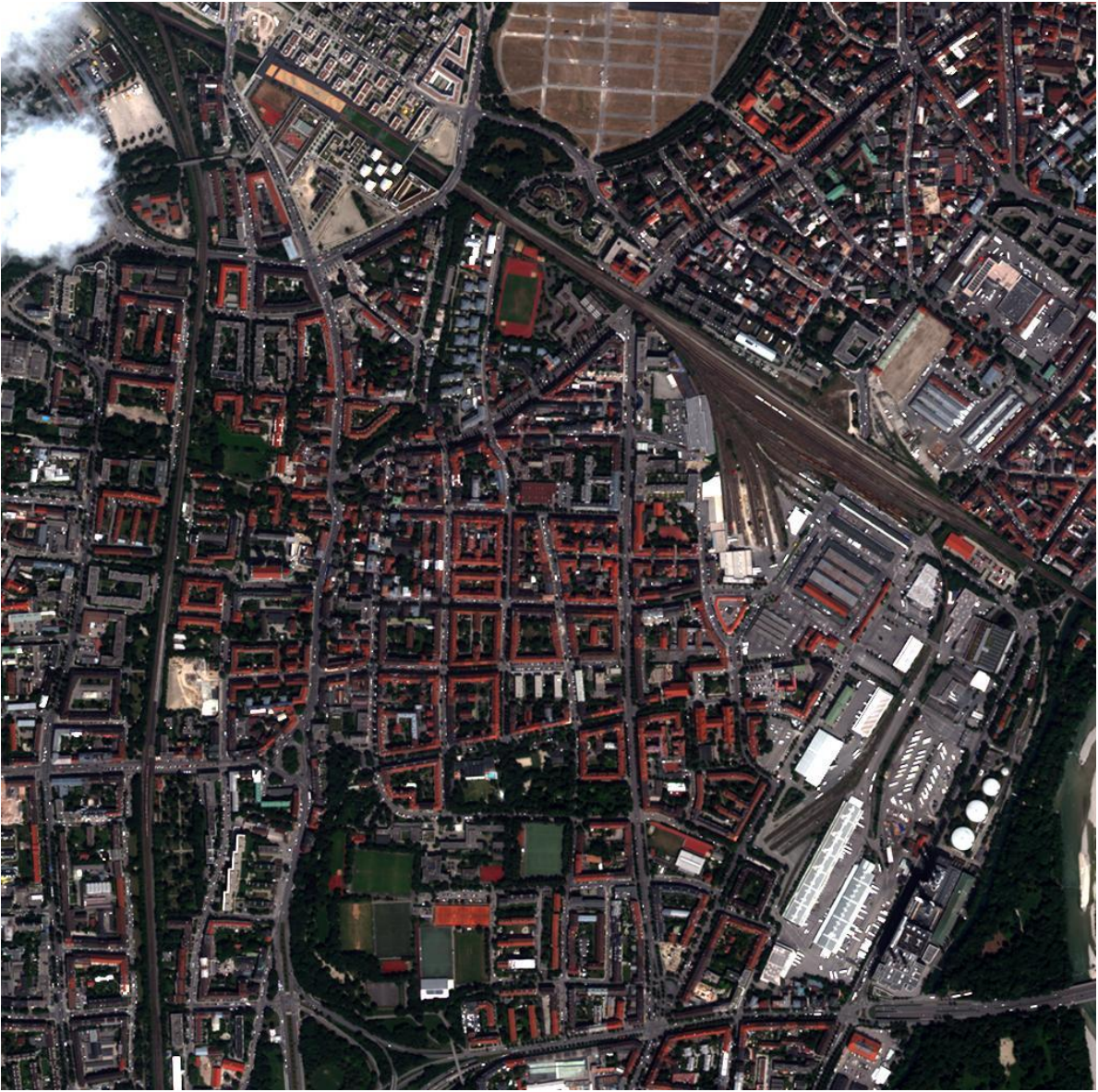

**Figure 3.** Multispectral image (bands: 5, 3, 2, size: 1024x1024) in original resolution for WV-2 ROI2.

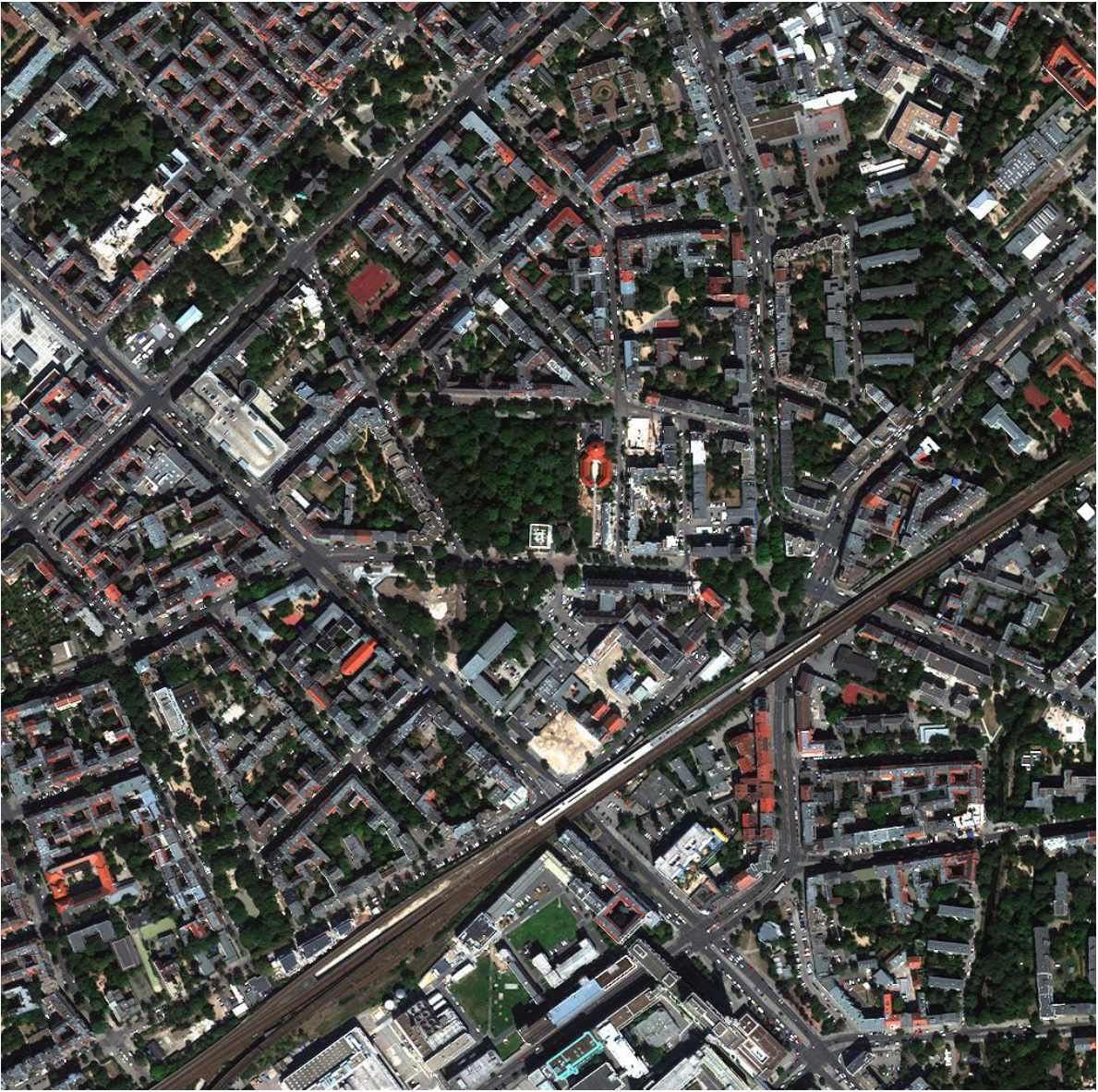

**Figure 4.** Multispectral image (bands: 3, 2, 1, size: 1024x1024) in original resolution for WV-4.

*6.2. Resolution*

As already discussed above, validation of pansharpening can be performed in the reduced scale (in this case the reference image exists, that is a low resolution MS image) or in the original or full resolution (reference not available in this case).

For reduced scale approach the question is which scale shall be used: 1:2 or 1:4 or even lower? In Figure 5. the visual appearance of such scale reduction is demonstrated for 1:2 and 1:4 scales. It can be seen, that images in scale 1:2 are still quite similar to original scale, whereas in scale 1:4 images differ significantly, and thus are less suitable for validation purposes. It is quite difficult to believe, that comparison conclusions derived in scale 1:4 will hold also for original resolution as it is performed practically in each comparison study.

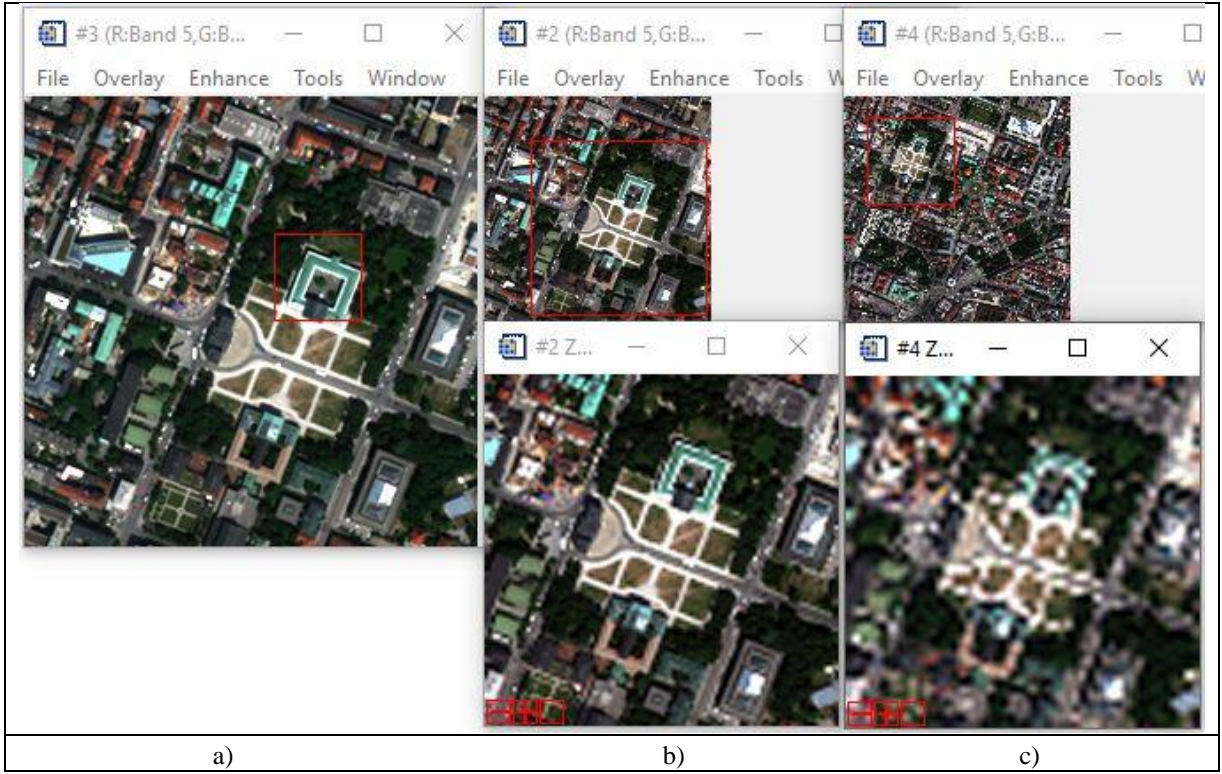

|      a)      |      b)      |      c)      |

**Figure 5.** WV-2 multispectral image (bands: 5, 3, 2, size: 256x256) in original resolution of 2 m (a). Low pass filtered MS image in scale 1:2 resolution of 4 m (b, top) and in scale 1:4 resolution of 8 m (c, top). Red boxes shown in (b, c, top) present zooms by factor 2 (b, bottom) and by factor 4 (c, bottom) using bicubic interpolation to produce original resolution of 2 m.

In this paper three resolutions for pansharpening validation are used:
- original resolution - pansharpening is performed in original resolution, then low pass filtered result (resolution 1:4) is compared to original MS image (consistency property [4, 37]),
- reduced resolution 1:4 - pansharpening is performed in reduced resolution 1:4 and result is compared to original MS image (synthesis property at reduced scale [4])
- reduced resolution 1:2 - pansharpening is performed in reduced resolution 1:2 and then low pass filtered result (resolution 1:4) is compared to original MS image (new proposal).

*6.3. Quality measures*

There exist a lot of quality measures, e.g. see overviews in [3, 29-33]. In case when the reference is known, then most measures are based on squared difference. Thus, for the relative comparison of methods it is enough the usage of very few measures. When the reference image is not available (what is a nominal case in pansharpening), then available/existing quality measures, e.g. such as Quality with No Reference (QNR) [34] are still not satisfactory, thus they are not considered in this paper.

In reduced resolution case the reference is known, thus two reference-based quality measures for validation are used: ERGAS and SAM.

ERGAS is defined as [39]

$$ERGAS(S_{hr,k}, S_{hr,k}^{ref}) = \frac{100}{scale} \sqrt{1/K \cdot \sum_{1}^{K} \left(\frac{RMSE(S_{hr,k}, S_{hr,k}^{ref})}{\mu(S_{hr,k}^{ref})}\right)^2} \qquad (25)$$

where

$$RMSE(S_{hr,k}, S_{hr,k}^{ref}) = \sqrt{1/J \cdot \sum_{1}^{J} \left(s_{hr,k,j} - s_{hr,k,j}^{ref}\right)^2} \qquad (26)$$

and

$$\mu(S_{hr,k}^{ref}) = 1/J \cdot \sum_{1}^{J} s_{hr,k,j}^{ref} \tag{27}$$

SAM is defined as [40]

$$SAM(S_{hr,k}, S_{hr,k}^{ref}) = \frac{1}{J} \sum_{1}^{J} \cos^{-1}\left( \sum_{1}^{K}(s_{hr,k,j} \cdot s_{hr,k,j}^{ref}) \Big/ \sqrt{\sum_{1}^{K} s_{hr,k,j}^{2} \cdot \sum_{1}^{K} s_{hr,k,j}^{ref\ 2}} \right) \tag{28}$$

## 7. Results

In this section nominal and enhanced versions (mostly only one correction applied) of CS and HPF based methods are compared on real satellite remote sensing data.

Following corrections are investigated:
- Haze correction (HC)
- Pan histogram matching (PHM)
- Weights estimation: use updated weights for intensity calculation (WE)
- Pan correction: use updated weights for intensity calculation and perform pan correction including pan histogram matching (WE + PC)
- Fusion result histogram matching (MHM)

The results are presented in the following Tables 3-5.

**Table 3.** Comparison of quality measures ERGAS (upper value) and SAM (lower value) for various pansharpening methods using different correction types: no correction (NC), haze correction (HC), pan histogram matching (PHM), weights estimation (WE), pan correction (PC) including weights estimation and fusion result multispectral histogram matching (MHM) for WV-2 ROI1[1].

| Resolution | Method | NC | HC | PHM | WE | WE +PC | MHM |
|---|---|---|---|---|---|---|---|
| Original | CS a | 2.05 | na | 1.38 | 1.49 | 0.89 | 1.39 |
|  |  | 0.90 | na | 0.66 | 0.65 | 0.52 | 0.92 |
|  | CS m | 2.33 | 2.24 | 1.58 | 1.57 | 1.01 | 1.56 |
|  |  | 0.38 | 0.81 | 0.38 | 0.38 | 0.38 | 0.75 |
|  | HPF a | 0.59 | na | 0.43 | na | 0.38 | 0.49 |
|  |  | 0.82 | na | 0.34 | na | 0.33 | 0.36 |
|  | HPF m | 0.48 | 0.50 | 0.49 | na | 2.57 | 0.52 |
|  |  | 0.35 | 0.32 | 0.30 | na | 0.30 | 0.31 |
| Reduced 4x4 | CS a | 3.26 | na | 2.97 | 2.96 | 2.98 | 2.83 |
|  |  | 4.46 | na | 4.35 | 4.43 | 4.38 | 4.36 |
|  | CS m | 3.30 | 3.09 | 2.86 | 2.82 | 2.83 | 2.79 |
|  |  | 4.07 | 3.75 | 4.07 | 4.07 | 4.07 | 4.23 |
|  | HPF a | 3.20 | na | 3.20 | na | 3.19 | 3.19 |
|  |  | 4.31 | na | 4.24 | na | 4.26 | 4.25 |
|  | HPF m | 3.06 | 2.95 | 3.06 | na | 3.09 | 3.08 |
|  |  | 4.08 | 3.72 | 4.08 | na | 4.08 | 4.10 |
| Reduced 2x2 | CS a | 2.23 | na | 1.98 | 1.78 | 1.92 | 1.92 |
|  |  | 2.52 | na | 2.40 | 2.40 | 2.40 | 2.54 |
|  | CS m | 2.26 | 2.21 | 1.94 | 1.68 | 1.85 | 1.89 |
|  |  | 2.30 | 2.39 | 2.30 | 2.31 | 2.30 | 2.50 |
|  | HPF a | 1.94 | na | 2.00 | na | 1.97 | 1.98 |
|  |  | 2.27 | na | 2.27 | na | 2.27 | 2.29 |
|  | HPF m | 1.91 | 1.90 | 1.94 | na | 1.93 | 1.95 |
|  |  | 2.23 | 2.10 | 2.23 | na | 2.24 | 2.26 |

[1] na stands for not available/applicable. Correction quality ranking: 1 (best), 2 (second best), correction is worse than without a correction.

**Table 4.** Comparison of quality measures ERGAS (upper value) and SAM (lower value) for various pansharpening methods using different corrections: no correction (NC), haze correction (HC), pan histogram matching (PHM), weights estimation (WE), pan correction (PC) including weights estimation and fusion result multispectral histogram matching (MHM) for WV-2 ROI2[1].

| Resolution | Method | NC | HC | PHM | WE | WE + PC | MHM |
|---|---|---|---|---|---|---|---|
| Original | CS a | 2.17 | na | 1.46 | 1.54 | 0.95 | 1.48 |
| | | 0.91 | na | 0.67 | 0.66 | 0.54 | 0.91 |
| | CS m | 2.51 | 2.42 | 1.69 | 1.66 | 1.11 | 1.68 |
| | | 0.41 | 0.86 | 0.41 | 0.41 | 0.41 | 0.68 |
| | HPF a | 0.50 | na | 0.46 | na | 0.41 | 0.51 |
| | | 0.36 | na | 0.36 | na | 0.35 | 0.38 |
| | HPF m | 0.54 | 0.53 | 0.52 | na | 0.59 | 0.55 |
| | | 0.32 | 0.34 | 0.32 | na | 0.32 | 0.33 |
| Reduced 4x4 | CS a | 3.41 | na | 3.11 | 3.08 | 3.12 | 2.99 |
| | | 4.68 | na | 4.59 | 4.66 | 4.61 | 4.53 |
| | CS m | 3.46 | 3.24 | 2.98 | 2.92 | 2.95 | 2.91 |
| | | 4.33 | 3.79 | 4.33 | 4.33 | 4.33 | 4.33 |
| | HPF a | 3.38 | na | 3.39 | na | 3.38 | 3.37 |
| | | 4.55 | na | 4.49 | na | 4.51 | 4.46 |
| | HPF m | 3.22 | 3.09 | 3.23 | na | 3.23 | 3.24 |
| | | 4.34 | 3.83 | 4.34 | na | 4.34 | 4.29 |
| Reduced 2x2 | CS a | 2.38 | na | 2.10 | 1.90 | 2.04 | 2.06 |
| | | 2.64 | na | 2.53 | 2.52 | 2.52 | 2.63 |
| | CS m | 2.44 | 2.38 | 2.05 | 1.80 | 1.97 | 2.02 |
| | | 2.44 | 2.48 | 2.44 | 2.44 | 2.44 | 2.55 |
| | HPF a | 2.06 | na | 2.11 | na | 2.09 | 2.10 |
| | | 2.40 | na | 2.40 | na | 2.40 | 2.41 |
| | HPF m | 2.03 | 2.02 | 2.05 | na | 2.37 | 2.07 |
| | | 2.37 | 2.22 | 2.37 | na | 2.37 | 2.38 |

[1] na stands for not available/applicable. Correction quality ranking: 1 (best), 2 (second best), correction is worse than without a correction.

**Table 5.** Comparison of quality measures ERGAS (upper value) and SAM (lower value) for various pansharpening methods using different corrections: no correction (NC), haze correction (HC), pan histogram matching (PHM), weights estimation (WE), pan correction (PC) including weights estimation and fusion result multispectral histogram matching (MHM) for WV-4[1].

| Resolution | Method | NC | HC | PHM | WE | WE + PC | MHM |
|---|---|---|---|---|---|---|---|
| Original | CS a | 2.08 | na | 2.14 | 2.08 | 1.60 | 2.12 |
|  |  | 1.23 | na | 1.20 | 1.22 | 1.12 | 1.30 |
|  | CS m | 2.75 | **23.12** | 2.81 | 2.56 | 2.79 | 3.36 |
|  |  | 0.96 | 1.74 | 0.96 | 0.96 | 0.96 | 2.70 |
|  | HPF a | 0.82 | na | 0.79 | na | 0.72 | 0.85 |
|  |  | 0.68 | na | 0.67 | na | 0.66 | 0.73 |
|  | HPF m | 1.00 | 1.09 | 0.94 | na | 2.05 | 1.04 |
|  |  | 0.63 | 1.71 | 0.63 | na | 0.63 | 0.68 |
| Reduced 4x4 | CS a | 4.36 | na | 4.66 | 4.34 | 4.59 | 4.48 |
|  |  | 5.79 | na | 5.67 | 5.76 | 5.69 | 5.59 |
|  | CS m | 3.96 | 3.68 | 4.28 | 3.93 | 4.21 | 4.10 |
|  |  | 5.44 | 4.52 | 5.44 | 5.44 | 5.44 | 5.23 |
|  | HPF a | 4.87 | na | 5.07 | na | 5.01 | 4.90 |
|  |  | 5.63 | na | 5.57 | na | 5.59 | 5.58 |
|  | HPF m | 4.60 | 4.37 | 4.79 | na | 4.73 | 4.64 |
|  |  | 5.44 | 4.71 | 5.44 | na | 5.44 | 5.32 |
| Reduced 2x2 | CS a | 3.92 | na | 4.12 | 3.92 | 4.04 | 4.04 |
|  |  | 3.81 | na | 3.79 | 3.79 | 3.79 | 3.83 |
|  | CS m | 3.78 | 3.71 | 4.04 | 3.78 | 3.94 | 3.98 |
|  |  | 3.73 | 3.56 | 3.72 | 3.73 | 3.72 | 3.79 |
|  | HPF a | 3.95 | na | 4.01 | na | 3.99 | 3.98 |
|  |  | 3.63 | na | 3.63 | na | 3.63 | 3.66 |
|  | HPF m | 3.88 | 3.81 | 3.93 | na | 3.94 | 3.92 |
|  |  | 3.60 | 3.38 | 3.60 | na | 3.60 | 3.60 |

[1] na stands for not available/applicable. Correction quality ranking: 1 (best), 2 (second best), correction is worse than without a correction.

Analysis of quality measure values may allow different questions to be answered, e.g. which method is the best one, which correction type brings most of the improvement or which is the most stable improvement, how consistent are the comparison results through different resolutions or data/sensors? The last question concerning the stability of results over different scales is very important because of the missing reference in pansharpening validation. If the results over different resolutions are not consistent, the proposed approach of validation in the reduced scale becomes questionable.

Correlation analysis (coefficients calculated from Tables 3-5) of quality values between different resolutions for various correction types is presented in Table 6.

**Table 6.** Comparison of correlation coefficient between pansharpening quality values of different pairs of resolutions: original resolution and reduced resolution 1:2 (upper value), original resolution and reduced resolution 1:4 (middle value) and reduced resolution 1:2 and reduced resolution 1:4 (lower value) for three datasets using different corrections: no correction (NC), haze correction (HC), pan histogram matching (PHM), weights estimation (WE), pan correction (PC) including weights estimation and fusion result multispectral histogram matching (MHM).

| Dataset | NC | HC | PHM | WE | WE + PC | MHM |
|---|---|---|---|---|---|---|
| WV-2 ROI1 | 0.23 | 0.35 | -0.54 | -0.96 | -0.55 | -0.35 |
|  | -0.37 | -0.41 | -0.66 | -0.93 | -0.56 | -0.56 |
|  | 0.78 | 0.63 | 0.98 | 1.00 | 0.99 | 0.95 |
| WV-2 ROI2 | 0.27 | 0.49 | -0.55 | -0.97 | -0.72 | -0.43 |
|  | -0.46 | -0.40 | -0.68 | -0.95 | -0.74 | -0.63 |
|  | 0.72 | 0.52 | 0.98 | 1.00 | 0.85 | 0.95 |
| WV-4 | 0.34 | 0.31 | 0.70 | 0.47 | 0.59 | 0.47 |
|  | -0.87 | -0.94 | -0.79 | -0.95 | -0.85 | -0.60 |
|  | -0.66 | -0.61 | -0.83 | -0.44 | -0.78 | -0.78 |

Analysis of results in Table 6 allows to derive following conclusions. There is no significant positive correlation between quality values between original scale and each of reduced scale. For WV-2 data quite strong correlation is observed between both reduced scales, whereas for WV-4 no positive correlation can be observed. It seems, that these results are not dependent on the type of pansharpening correction type. Off course these observations need to be verified in further/larger studies. Nevertheless, the reduced pansharpening validation approach seems to be not reliable/stable/appropriate and cannot be unconditionally recommended for pansharpening methods validation/comparison. Thus, such large-scale studies as e.g. conducted in [46] should be evaluated/taken with a great care.

## 8. Discussion

Pansharpening methods produce MS image in the high resolution of panchromatic image (this resolution is called original resolution throughout this paper). Unfortunately, the reference of MS image is missing in this original resolution making the validation of a pansharpening method difficult. Experimental study in this paper has shown that validation approaches of pansharpening methods based on consistency property and reduced scale do not deliver results which correlate over scales thus causing doubts of being suitable for evaluation/comparison purposes. It seems at the moment, that the only reasonable validation approach is to perform comparison in a particular application with the usage of their ground truth data.

## 9. Conclusions

Experimental study conducted in this paper has shown that existing validation approach in reduced resolution and/or consistency check should be applied very carefully for pansharpening method validation/comparison.

Presented methodology is not limited only to the pansharpening. It can be easily applied to the hyperspectral and multispectral sharpening too.

The future work could be directed towards devepolment of pansharpening validation approaches free of drawbacks discussed above.

**Acknowledgments:** I would like to thank DigitalGlobe and European Space Imaging (EUSI) for the collection and provision of WorldView-2 scene over the Munich city and WorldView-4 scene over the Berlin city.

**References**

1. Pohl, C., van Genderen, J. Review article Multisensor image fusion in remote sensing: concepts, methods and applications. *International Journal of Remote Sensing* **1998**, *19 (5)*, 823–854.
2. Wang, Z., Ziou, D., Armenakis, C., Li, D., Li, Q. A comparative analysis of image fusion methods. *IEEE Transactions on Geoscience and Remote Sensing* **2005**, *43 (6)*, 1391-1402.

3. Alparone, L., Wald, L., Chanussot, J., Thomas, C., Gamba, P., Bruce, L.M. Comparison of pansharpening algorithms: Outcome of the 2006 GRS-S data-fusion contest. *IEEE Transactions on Geoscience and Remote Sensing* **2007**, *45 (10)*, 3012-3021.
4. Thomas, C., Ranchin, T., Wald, L., Chanussot, J. Synthesis of multispectral images to high spatial resolution: a critical review of fusion methods based on remote sensing physics. *IEEE Transactions on Geoscience and Remote Sensing* **2008**, *46 (5)*, 1301–1312.
5. Ehlers, M., Klonus, S., Åstrand, P.J., Rosso, P. Multisensor image fusion for pansharpening in remote sensing. *International Journal of Image and Data Fusion* **2010**, *1 (1)*, 25-45.
6. Amro, I., Mateos, J., Vega, M., Molina, R., Katsaggelos, A. A survey of classical methods and new trends in pansharpening of multispectral images. *EURASIP J. Adv. Signal Process*. **2011**, *79*, 1–22.
7. Alparone, l., Aiazzi, B., Baronti, S., Garzelli, A. *Remote Sensing Image Fusion*, CRC Press: Boca Raton, FL, USA, 2015.
8. Pohl, C., van Genderen, J. Structuring contemporary remote sensing image fusion. *International Journal of Image and Data Fusion* **2015**, *6 (1)*, 3–21.
9. Vivone, G., Alparone, L., Chanussot, J., Dalla Mura, M., Garzelli, A., Licciardi, G.A., Restaino, R., Wald, L. A critical comparison among pansharpening algorithms. *IEEE Transactions on Geoscience and Remote Sensing* **2015**, *53 (5)*, 2565-2586.
10. Pohl, C., van Genderen, J. Remote Sensing Image Fusion: A Practical Guide, CRC Press: Boca Raton, FL, USA, 2017.
11. Meng, X., Shen, H., Li, H., Zhang, L., Fu, R. Review of the pansharpening methods for remote sensing images based on the idea of meta-analysis: Practical discussion and challenges. *Information Fusion* **2019**, *46*, 102–113.
12. Vivone, G., Dalla Mura, M., Garzelli, A., Restaino, R., Scarpa, G., Ulfarsson, M.O., Alparone, L., Chanussot, J. A New Benchmark Based on Recent Advances in Multispectral Pansharpening: Revisiting pansharpening with classical and emerging pansharpening methods. *IEEE Geoscience and Remote Sensing Magazine* **2020**, doi: 10.1109/MGRS.2020.3019315
13. Aanæs, H., Sveinsson, J.R., Nielsen, A.A., Bøvith, T., Benediktsson, J.A. Model-based satellite image fusion. *IEEE Transactions on Geoscience and Remote Sensing* **2008**, *46 (5)*, 1336-1346.
14. Zhukov, B., Oertel, D., Lanzl, F., Reinhäckel, G. Unmixing-based multisensor multiresolution image fusion. *IEEE Transactions on Geoscience and Remote Sensing* **1999**, *37 (3)*, 1212-1226.
15. Palubinskas, G., Reinartz, P. Multi-resolution, multi-sensor image fusion: general fusion framework. Proc. of Joint Urban Remote Sensing Event (JURSE), Munich, Germany, 30 March-1 April, 2011; Stilla, U., Gamba, P., Juergens, C., Maktav, D., Eds.; IEEE: Piscataway, NJ, USA, 2011.
16. Palubinskas, G. Fast, simple and good pan-sharpening method. *Journal of Applied Remote Sensing* **2013**, *7 (1)*, 1-12.
17. Palubinskas, G. Model-based view at multi-resolution image fusion methods and quality assessment measures. *International Journal of Image and Data Fusion* **2016**, *7(3)*, 203-218, doi: 10.1080/19479832.2016.1180326
18. Palubinskas, G. Pan-sharpening approaches based on unmixing of multispectral remote sensing imagery. In The International Archives of the Photogrammetry, Remote Sensing and Spatial Information Sciences, Proc. XXIII ISPRS Congress, Prague, Czech Republic, 12-19 July, 2016; vol. XLI-B7, 693-702. doi:10.5194/isprs-archives-XLI-B7-693-2016
19. Li, S., and Yang, B. A new pan-sharpening method using a compressed sensing technique. *IEEE Trans. Geosci. Remote Sens.* **2011**, *49 (2)*, 738–746.
20. Alparone, L., Garzelli, A., Vivone, G. Intersensor statistical matching for pansharpening: theoretical issues and practical solutions. *IEEE Transactions on Geoscience and Remote Sensing* **2017**, *55 (8)*, 4682-4695.
21. Otazu, X., González-Audicana, M., Fors, O., Nunez, J. Introduction of sensor spectral response into image fusion methods. Application to wavelet-based methods. *IEEE Transactions on Geoscience and Remote Sensing* **2005**, *43 (10)*, 2376-2385.
22. González-Audícana, M., Otazu, X., Fors, O., Alvarez-Mozos, J. A Low Computational-Cost Method to Fuse IKONOS Images Using the Spectral Response Function of Its Sensors. *IEEE Transactions on Geoscience and Remote Sensing* **2006**, *44 (6)*, 1683-1691.
23. Švab, A., Oštir, K. High-Resolution Image Fusion: Methods to Preserve Spectral and Spatial Resolution. *Photogrammetric Engineering & Remote Sensing* **2006**, *72 (5)*, 565–572.
24. Aiazzi, B., Baronti, S., and Selva, M. Improving component substitution pansharpening through multivariate regression of MS+Pan data. *IEEE Trans. Geosci. Remote Sens.* **2007**, *45*, 3230–3239.
25. Kim, Y.H., Eo, Y.D., Kim, Y.S., Kim, Y.I. Generalized IHS-Based Satellite Imagery Fusion Using Spectral Response Functions. *ETRI Journal* **2011**, *33 (4)*, 497-505.

26. Palubinskas, G. Model-based image adjustment for a successful pan-sharpening, arXiv:2103.03062 [eess.IV], 2021, 17 pages.
27. Lolli, S., Alparone, L., Garzelli, A., Vivone, G. Haze correction for contrast-based multispectral pansharpening, *IEEE Geoscience and Remote Sensing Letters* **2017**, 14(12), 2255-2259.
28. Aiazzi, A., Alparone, L., Baronti, S., Garzelli, A., Selva, M. MTF-tailored Multiscale Fusion of High-resolution MS and Pan Imagery. *Photogrammetric Engineering & Remote Sensing* **2006**, 72 (5), 591–596.
29. Garzelli, A., Aiazzi, B., Alparone, L., Lolli, S., Vivone, G. Multispectral pansharpening with radiative transfer-based detail-injection modeling for preserving changes in vegetation cover. *Remote Sensing* **2018**, 10, 1308.
30. Wald, L., Ranchin, T., Mangolini, M. Fusion of satellite images of different spatial resolutions: Assessing the quality of resulting images. *Photogramm. Eng. Remote Sens.* **1997**, *63 (6)*, 691–699.
31. Aiazzi, B., Alparone, L., Baronti, S., Garzelli, A. Quality assessment of pansharpening methods and products. *IEEE Geosci. Remote Sens. Soc. Newsl.* **2011**, *161*, December, 10–18.
32. Makarau, A., Palubinskas, G., Reinartz, P. Analysis and selection of pan-sharpening assessment measures. *Journal of Applied Remote Sensing* **2012**, *6 (1)*, 1-20.
33. Palubinskas, G. Mystery behind similarity measures MSE and SSIM. Proc. International Conference on Image Processing (ICIP), Paris, France, 27-30 October, 2014; IEEE: Piscataway, NJ, USA: IEEE, 575-579.
34. Palubinskas, G. Image similarity/distance measures: what is really behind MSE and SSIM? *International Journal of Image and Data Fusion* **2017**, *8 (1)*, 32-53. doi:10.1080/19479832.2016.1273259
35. Palubinskas, G. Joint quality measure for evaluation of pansharpening accuracy. *Remote Sensing* **2015**, *7 (7)*, 9292-9310.
36. Alparone, L., Aiazzi, B., Baronti, S., Garzelli, A., Nencini, F., Selva, M. Multispectral and panchromatic data fusion assessment without reference. *Photogramm. Eng. Remote Sens.* **2008**, *74 (2)*, 193–200.
37. Palsson, F., Sveinsson, J.R., Ulfarsson, M.O., Benediktsson, J.A. Quantitative Quality Evaluation of Pansharpened Imagery: Consistency Versus Synthesis. *IEEE Transactions on Geoscience and Remote Sensing* **2016**, 54 (3), 1247-1259.
38. Vivone, G., Alparone, L., Garzelli, A., Lolli, S. Fast reproducible pansharpening based on instrument and acquisition modeling: AWLP revisited. *Remote Sensing* **2019**, 11, 2315, 1-23.
39. Wald, L. Quality of high resolution synthesized images: Is there a simple criterion. Proc. 3rd Conf. Fusion Earth Data, Ranchin, T. and Wald, L. Eds., 2000, pp. 99–105.
40. R. H. Yuhas, A. F. H. Goetz, and J. W. Boardman, "Discrimination among semi-arid landscape endmembers using the Spectral Angle Mapper (SAM) algorithm in Proceeding Summaries 3rd Annual JPL Airborne Geoscience Workshop, 1992, pp. 147–149.
41. Spectral Response for DigitalGlobe Earth Imaging Instruments, Available online: https://dg-cms-uploads-production.s3.amazonaws.com/uploads/document/file/105/DigitalGlobe_Spectral_Response_1.pdf (accessed on 30 November 2020)
42. Stark, P.B., Parker, R.L. Bounded-variable least-squares: an algorithm and applications. *Computational Statistics* **1995**, *10*, 129-141.
43. Borg, E., Fichtelmann, B., Richter, R., Bachmann, M., 2006. Verfeinerung der örtlichen Auflösung multispektraler Fernerkundungsdaten, Patent DE 10 2004 039 634 A1 2006.02.23.
44. Makarau, A., Palubinskas, G., Reinartz, P. Multiresolution image fusion: Phase congruency for spatial consistency assessment. Proc. ISPRS Technical Commision VII Symposium - 100 Years ISPRS, IAPRS, 2010, pp. 383-388.
45. Makarau, A., Palubinskas, G., Reinartz, P. Multi-sensor data fusion for urban area classification. Proc. Joint Urban Remote Sensing Event, IEEE, 2011, pp. 21-24.
46. Meng, X., Xiong, Y., Shao, F., Shen, H., Sun, W., Yang, G., Yuan, Q., Fu, R., Zhang, H. A Large-Scale Benchmark Data Set for Evaluating Pansharpening Performance: Overview and implementation. *IEEE Geoscience and Remote Sensing Magazine*, March 2021, pp. 18-52.